\documentclass[10pt,twocolumn,letterpaper]{article}

\usepackage{iccv}              % To produce the CAMERA-READY version
\usepackage{multirow}
\usepackage{multicol}
\usepackage{caption}

\newcommand{\ours}{GigaTok}

\definecolor{iccvblue}{rgb}{0.21,0.49,0.74}
\usepackage[pagebackref,breaklinks,colorlinks,allcolors=iccvblue]{hyperref}
\usepackage{graphicx}
\usepackage{lipsum}
\usepackage{hyperref}
\usepackage[accsupp]{axessibility}  % Improves PDF readability for those with disabilities.

\makeatletter
\def\blfootnote{\xdef\@thefnmark{}\@footnotetext}
\renewcommand{\@makefntext}[1]{\noindent#1} % This removes the indent
\makeatother

\def\paperID{12572} % *** Enter the Paper ID here
\def\confName{ICCV}
\def\confYear{2025}

\title{GigaTok: Scaling Visual Tokenizers to 3 Billion Parameters \\for Autoregressive Image Generation}

\author{Tianwei Xiong$^{1}$
\and
Jun Hao Liew$^{2}$
\and
Zilong Huang$^{2}$
\and
Jiashi Feng$^{2}$
\and
Xihui Liu$^{1\dagger}$
\and
$^1$The University of Hong Kong
\and
$^2$ByteDance Seed
\and
Project page: \href{https://silentview.github.io/GigaTok/}{https://silentview.github.io/GigaTok/}
}

\begin{document}

\maketitle
% \blfootnote{\noindent$^*$ Work partly done as an Intern at ByteDance. $^\dagger$ Correspondence Author.}

\blfootnote{$^\dagger$ Corresponding Author.}

\begin{abstract}

In autoregressive (AR) image generation, visual tokenizers compress images into compact discrete latent tokens, enabling efficient training of downstream autoregressive models for visual generation via next-token prediction. While scaling visual tokenizers improves image reconstruction quality, it often degrades downstream generation quality—a challenge not adequately addressed in existing literature. To address this, we introduce GigaTok, the first approach to simultaneously improve image reconstruction, generation, and representation learning when scaling visual tokenizers. We identify the growing complexity of latent space as the key factor behind the reconstruction \vs generation dilemma. To mitigate this, we propose semantic regularization, which aligns tokenizer features with semantically consistent features from a pre-trained visual encoder. This constraint prevents excessive latent space complexity during scaling, yielding consistent improvements in both reconstruction and downstream autoregressive generation. Building on semantic regularization, we explore three key practices for scaling tokenizers:~(1) using 1D tokenizers for better scalability,~(2) prioritizing decoder scaling when expanding both encoder and decoder, and ~(3) employing entropy loss to stabilize training for billion-scale tokenizers. By scaling to \textbf{3 billion} parameters, GigaTok achieves state-of-the-art performance in reconstruction, downstream AR generation, and downstream AR representation quality.  
\end{abstract}

\begin{figure}
    \centering
\includegraphics[width=\linewidth]{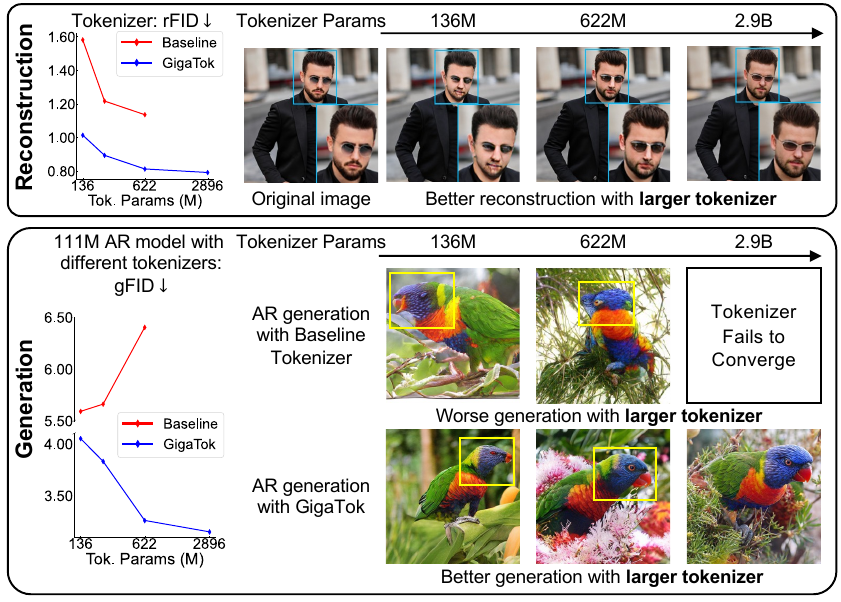}
    \caption{\textbf{Reconstruction \vs generation dilemma}: Naively scaling visual tokenizers achieves better reconstruction but degrades downstream autoregressive (AR) generation. In contrast, \ours{} achieves better performance for both reconstruction and generation as tokenizers scale up.}
    \vspace{-10pt}
    \label{fig:teaser}
\end{figure}

%%%%%%%%% ABSTRACT

\begin{figure*}[t]
% \vspace{-0.1in}
    \centering
    \includegraphics[width=\linewidth]{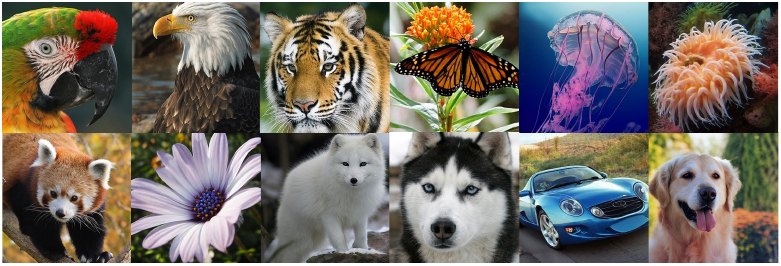}
    % \vspace{-0.30in}
    \caption{
    The 2.9B \ours{} achieves SOTA autoregressive image generation with a 1.4B AR model on ImageNet 256$\times$256 resolution.
    }
    % \vspace{-0.3in}
    \label{fig:qual_grid}
\end{figure*}

\section{Introduction}
Autoregressive (AR) language models (LM) have emerged as a promising approach for visual generation~\cite{llamagen, magvit-v2, vqgan, vit_vqgan}, driven by their proven scalability~\cite{llama, llama2, llama3, gemini, gemini_1_5, gpt4, deepseek_llm, deepseek-r1, deepseek-v3} and the potential for unified multimodal modeling~\cite{janus, janus-pro, qu2024tokenflow}. The AR image generation framework consists of a visual tokenizer and a downstream AR generator. The tokenizer encodes images into discrete tokens,  trained with image reconstruction supervision, while the AR generator models the distribution of these discrete tokens through next-token prediction. 
The image tokenizer plays a pivotal role in AR visual generation,  providing a compact and expressive latent space that enables effective generative modeling by downstream AR models.

Despite its pivotal role, scaling of visual tokenizer is rarely explored in the literature. In fact, unlike the downstream AR models whose scalability has been widely validated~\cite{janus, janus-pro, wang2024emu3, videopoet}, scaling the visual tokenizer presents a significant challenge. Specifically, 
there exists a  \textit{reconstruction \vs generation dilemma}, where scaling tokenizer improves reconstruction fidelity but degrades downstream generation quality, as shown in Fig.~\ref{fig:teaser}. This dilemma is also observed in prior works~\cite{vit,scale_tokenizers}. 
In this work, we seek to overcome this limitation and explore strategies for effectively scaling tokenizers to enhance both reconstruction and generation performance.

To investigate the root cause of this dilemma, we propose an AR probing scheme that trains a lightweight downstream generative AR model to monitor the tokenizer's training process. Surprisingly, we find that as tokenizers scale, the downstream AR model struggles more to learn the resulting token distribution, as evidenced by the increasing AR generation loss. This suggests that the larger tokenizers produce a more complex token space, making it increasingly difficult for AR models to learn effectively.

To address this challenge, we introduce pre-trained visual representation models (\eg DINOv2~\cite{dinov2}) to regularize tokenizers. 
Specifically, we leverage a \textit{semantic regularization} loss during tokenizer training, encouraging high similarity between tokenizer features and the pre-trained model features. Such regularization helps constrain the latent space complexity, preventing the tokenizer from learning overly complicated latent token dependencies that hinder downstream AR generative modeling. Moreover, we design a vector-quantized (VQ) tokenizer with a hybrid CNN-Transformer architecture as the backbone, suitable for both 1D and 2D tokenizers, and explore best practices for scaling tokenizers: (1) \textit{1D tokenizers} exhibit better scalability compared to 2D tokenizers; (2) \textit{Asymmetric model scaling},   prioritizing decoder scaling over encoder scaling, proves effective;  (3) \textit{Entropy loss}~\cite{magvit-v2} becomes crucial for convergence when training tokenizers with billion-level parameters. With our semantic regularization and three key scaling strategies, we effectively scale \ours{}~to 3 billion parameters, overcoming the reconstruction \vs generation dilemma.

We summarize our contributions as follows:
\begin{itemize}
  \item We identify that the reconstruction \vs generation dilemma in tokenizer scaling stems from increased latent space complexity in larger tokenizers. To address this, we propose semantic regularization, effectively mitigating the dilemma and enabling tokenizer scaling.
  \item We explore best practices for scaling tokenizers, including 1D tokenizers with hybrid CNN-Transformer architecture, asymmetric encoder-decoder scaling, and entropy loss for billion-scale tokenizers.
  \item Our \ours{} is the first tokenizer scaled to 3B, achieving state-of-the-art reconstruction, downstream AR generation, and downstream AR representation on ImageNet.
\end{itemize}

\section{Related Work}

% VQ, (RQ, LFQ, FSQ, IBQ), scaling
\noindent\textbf{Image tokenizers.}
Image tokenizers map image inputs into discrete~\cite{vqgan, vqvae, vit_vqgan} or continuous~\cite{vae} tokens which can be modeled by downstream generative models. For discrete tokenizers, Vector Quantization~(VQ)~\cite{vqgan, vqvae, vit_vqgan} is dominantly adopted. Recently, new quantization methods~\cite{magvit-v2, bsq, ibq_scale, scale_codebook} have also been proposed for better scaling of codebook size. However, how to properly scale up tokenizer models is insufficiently studied in existing literature. ViT-VQGAN~\cite{vit_vqgan} and TiTok~\cite{titok} utilize transformer architecture to enable convenient scaling of tokenizers, but end up training their best generative models on smaller tokenizer versions. A concurrent work, ViTok~\cite{scale_codebook}, suggests de-prioritizing VAE scaling due to its less predictable effect for downstream diffusion models. We observe a similar reconstruction \vs generation dilemma in scaling discrete tokenizers, and provide our analysis and solution to it.

% Continuous tokenizers are built upon Variational Autoencoders (VAE)~\cite{vae, vae_journal}, and discrete tokenizers such as VQ-GAN~\cite{vqgan} quantize visual features into discrete visual tokens during training. 

% In contrast, we provide a detailed analysis of the reconstruction \vs generation dilemma for scaling tokenizers and the solution to it.

% MaskGIT MaskBit, MAGVIT-v2 / Open-MAGVIT-v2, IBQ, VQ-GAN, ViT-VQGAN, LlamaGen / VAR, ImageFolder

\noindent\textbf{Autoregressive Visual Generation.} Autoregressive visual generative models~\cite{vit_vqgan, rq, llamagen, openmagvit-v2, ibq_scale, vqvae, lumina, wang2024emu3, larp} follow the next-token-prediction~(NTP) approach of LLMs, enabling the leverage of advancements in LLMs and simplifying the path to unified multi-modal generation. Other variants utilize visual-specific paradigms such as mask image modeling~\cite{chang2022maskgit, maskbit, titok, magvit-v2} and next-scale-prediction~\cite{var, imagefolder} for better performance. We reveal that scaling tokenizers helps NTP AR models to be comparable to these variants.

\noindent\textbf{Semantic Guidance for Visual Generative Models and Tokenizers.} The guidance from visual foundation models~\cite{dino, dinov2, CLIP, siglip, MAE} has been used to improve training convergence speed and quality~\cite{repa, lightningDiT} of visual generative models, as well as enhancing representation quality or downstream performance of visual tokenizers~\cite{imagefolder, softvq, scale_codebook, seed, wu2024vila, yu2023spae, vqct, lightningDiT, repa_in_vae, maetok, digit, vqkd, ma2025unitok}. REPA~\cite{repa} presents impressive performance improvements brought by a simple representation alignment strategy, and recently, VA-VAE~\cite{lightningDiT} shows the significant benefits of semantic guidance to the reconstruction-generation Pareto Frontier of VAEs. 
% As a convenient plug-and-play design, semantic regularization can be applied in a wide range of scenarios. 
Different from existing work, GigaTok novelly reveals the critical role of semantic regularization for resolving the reconstruction \vs generation dilemma in scaling visual tokenizers.

\section{Pilot Study }
\label{sec:method}

We first introduce AR Probing as a proxy to effectively monitor the tokenizer's effectiveness for downstream generation (Sec \ref{sec:AR Probe}), followed by a pilot experiment that investigates the reconstruction \vs generation challenges when naively scaling visual tokenizers (Sec \ref{sec:preliminary exp}).

\subsection{AR Probing for Tokenizer Evaluation
%: Monitoring Tokenizer Performance for Downstream Generation Models
}
\label{sec:AR Probe}

In autoregressive visual generation, the training of the tokenizer and downstream AR model are performed in separate stages. In the first stage, a visual tokenizer is trained to compress images into discrete tokens, 
optimized with reconstruction objective. In the second stage, the downstream generative model is trained based on the discrete tokens from the pre-trained tokenizer.
However, a tokenizer that performs well in terms of reconstruction fidelity in the first stage may not necessarily lead to better performance for downstream generative models. Thus, it is crucial to evaluate the effectiveness of the trained tokenizers for downstream generation alongside its reconstruction quality.

Despite its importance, assessing how a tokenizer influences downstream generation models can be computationally expensive. 
% For example, training a relatively small 343M parameter LlamaGen-L model for 300 epochs takes 170 hours on 64 V100 GPUs 
For example, sufficiently training a 343M parameter downstream AR generator takes 170 hours on 64 V100 GPUs.
To address this challenge, we introduce \textbf{AR Probing}, inspired by Linear Probing in representation learning literature~\cite{MAE, igpt}. The key idea is to use the performance of a small AR model as a proxy to reflect the performance trends of large-scale AR models. 

Specifically, we use the tokenizer to train a small Llama-style model~\cite{llama, llamagen}~(111M parameters) for 50 epochs, and evaluate its gFID~\cite{FID}, validation loss, and linear probing accuracy~\cite{igpt, MAE} 
% of this small model 
for a fair comparison between different tokenizers. Training the proposed AR Probing model for evaluating tokenizers is 10$\times$ more efficient than training the original 343M downstream AR model. Our experiments in Sec.~\ref{subsec:settings} (Fig.~\ref{fig:ar_probe_correlation}) demonstrate that the trends observed with AR Probing align with the performance of the large-scale AR models after sufficient training.

\noindent \textbf{gFID.} 
The generation FID~\cite{FID} of AR probing indicates the overall image generation performance of the two-stage framework. It reflects both the reconstruction fidelity of the tokenizer and how well the downstream AR probing model can learn the dependency of the visual tokens (\ie, learnability of the token distribution).

% \smallskip 
\noindent \textbf{Validation loss}. 
%As gFID provides an entangled metric for tokenizer reconstruction and latent space learnability, w
We use the validation loss of the AR probing model to measure the learnability of the latent tokens as a disentangled factor. The validation loss is calculated as an average of the token-wise cross-entropy loss in the next-token-prediction paradigm on ImageNet~\cite{russakovsky2015imagenet} 50k validation set. With the same vocabulary size, the same number and structure of visual tokens, and the same AR probing model, larger validation loss indicates a latent space that is more difficult for the AR model to learn. Therefore, we use validation loss to reflect the latent space complexity and learnability for AR models.

\noindent \textbf{Linear probing accuracy.}
Beyond visual generation quality, we also investigate whether scaling tokenizers will lead to better visual representations of AR models, which may provide inspiration for future research in unified multimodal understanding and generation with AR models. To assess the representation quality, we adopt the standard practice~\cite{igpt, MAE} of linear probing accuracy using features from the middle Transformer layer of the AR probing model.

\begin{figure}[]
% \vspace{-0.1in}
    \centering
    \includegraphics[width=\linewidth]{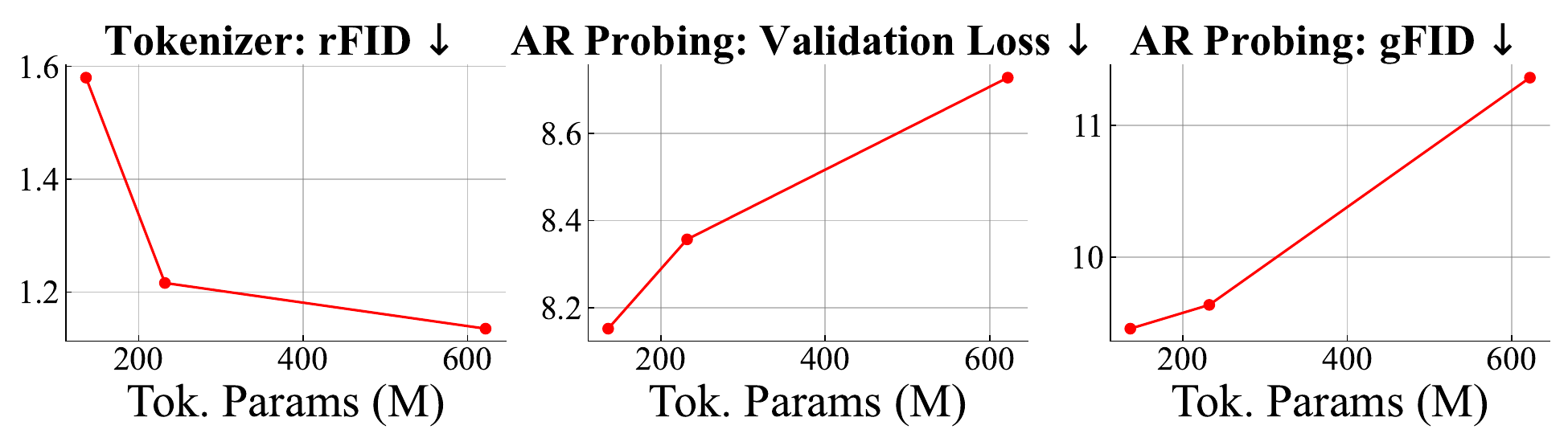}
    % \vspace{-0.30in}
    \caption{\textbf{Scaling trend for vanilla 1D tokenizers.} As the model size increases, the reconstruction quality of vanilla tokenizers improves but the downstream AR Probing gFID consistently degrades. The increasing AR Probing validation loss indicates that scaling vanilla tokenizers results in a more complex latent space, making it difficult for AR models to learn effectively. 
    }
    \vspace{-10pt}
    \label{fig:tok_size_scale_nodist}
\end{figure}

\subsection{
Naively Scaling Tokenizers Does Not Work
}
\label{sec:preliminary exp}

To study the challenges when naively scaling visual tokenizers,
we train three vector-quantized tokenizers\footnote{The tokenizer architectures are described in Sec.~\ref{sec:structures}}
on ImageNet~\cite{russakovsky2015imagenet} at 256$\times$256 resolution with increasing model sizes. 
As shown in Fig.~\ref{fig:tok_size_scale_nodist}, as the tokenizer size increases, although the reconstruction quality (rFID) consistently improves, the AR generation performance (gFID) significantly degrades. This highlights the reconstruction \vs generation dilemma in tokenizer scaling. 
Moreover, we observe that the validation loss of AR Probing consistently increases as the tokenizers scale, indicating that larger tokenizers lead to complicated token dependencies that are more difficult for the AR model to learn. This observation motivates us to design the semantic regularization to constrain the latent space complexity of the tokenizer and therefore break the reconstruction \vs generation dilemma in Sec.~\ref{sec:semantic regularization}.

\section{\ours{}
}

In this section, we introduce the model structure and training strategies for our scalable visual tokenizer, \ours{}. In Sec.~\ref{sec:structures}, we present a tokenizer backbone supporting 1D and 2D token structures, and discuss the asymmetric scaling strategies for the encoder and decoder. In Sec.~\ref{sec:semantic regularization}, we introduce semantic regularization, which breaks the reconstruction \vs generation dilemma by regularizing the complexity of the latent space with pre-trained visual representations. In Sec.~\ref{sec:entropy loss}, we show how entropy loss~\cite{magvit-v2} facilitates the convergence of billion-scale tokenizers.
\subsection{Architecture}
\label{sec:structures}

\begin{figure}[t]
% \vspace{-0.1in}
    \centering
    \includegraphics[width=\linewidth]{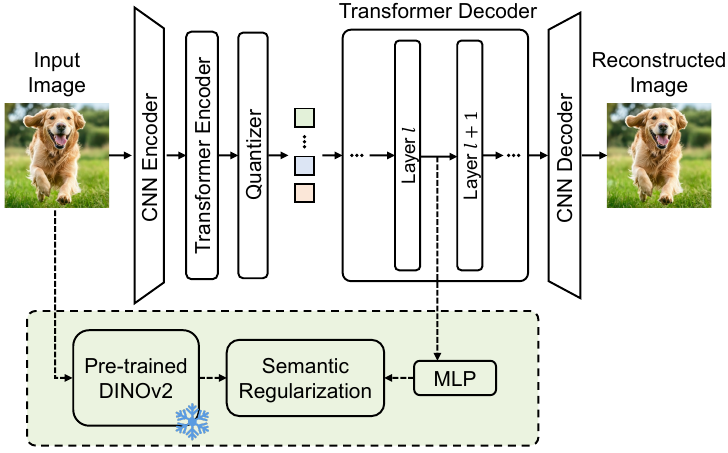}
    % \vspace{-0.30in}
    \caption{\textbf{\ours{} architecture and semantic regularization.} \textit{Top:} We use a hybrid CNN-Transformer design for our visual tokenizer. The transformer layers are implemented with ViT for 2D tokenizer and Q-Former for 1D tokenizer. \textit{Bottom:} We use a frozen DINOv2~\cite{dinov2} image encoder
    for semantic regularization.
    }
    \vspace{-5pt}
    \label{fig:tok_pipeline}
\end{figure}

The CNN~\cite{CNN} architectures have been the dominant choices for image tokenizers~\cite{vqgan, openmagvit-v2, magvit-v2, scale_codebook} due to their effectiveness in capturing fine-grained local details. Yet, Transformers are more scalable architectures with less inductive bias. Thus, we design a vector quantized tokenizer backbone with a hybrid architecture that combines CNN~\cite{CNN, vqgan} and Transformer~\cite{attention, vit, DETR} for encoder and decoder~(Fig.~\ref{fig:tok_pipeline}).
Specifically, our encoder consists of a series of CNN blocks that progressively downsamples the input image by a factor of $p$, followed by Transformer layers and a vector quantizer to produce discrete latent codes. 
Similarly, our decoder consists of multiple Transformer layers, followed by CNN decoders which upsamples the features to obtain the reconstructed image\footnote{Throughout this work, we use downsample ratio $p=16$, codebook dimension $D=8$, and codebook size $16384$ by default.}. Our tokenizer architecture can be adapted to both 1D and 2D tokenizers by using different Transformer designs introduced in the next two paragraphs.

\smallskip
\noindent \textbf{2D tokenizers with ViT.} For 2D tokenizers, the Transformers in both tokenizer encoder and decoder are implemented by ViT~\cite{vit} architecture. 2D structures of the latent features and tokens are preserved throughout the tokenizer. 

\smallskip
\noindent \textbf{1D tokenizers with Q-Former.}
For 1D tokenizers, we implement the Transformer modules in both encoder and decoder as Q-Formers~\cite{blip2, DETR}. The Q-Former in the encoder employs 1D queries, transforming 2D input features into 1D latent tokens. The Q-Former in the decoder utilizes 2D queries to transform 1D latent tokens back to 2D features, which are then passed to the CNN decoder to reconstruct images. The 1D tokenizers remove the 2D inductive bias and demonstrate better scalability than 2D tokenizers in our experiments (Sec.~\ref{sec:ablation}).

\smallskip
\noindent \textbf{Asymmetric encoder-decoder scaling.}
% The encoder and decoder components of the tokenizers serve different roles in image compression. 
Since the decoder faces the more challenging task of reconstructing images from lossy latent codes, we adopt an asymmetric design for more efficient parameter allocation. Specifically, we scale both the encoder and decoder, while ensuring that the decoders are always larger than the encoders. In practice, we maintain the same and fixed size for the CNN encoder/decoder and only increase the depth and width of the Transformer modules for scaling. %To summarize, we propose scaling both the encoder and decoder, with the decoders consistently being larger than the encoders.

\label{sec:semantic regularization}

\subsection{Semantic Regularization}
% \label{subsec:semantic-reg}
\label{sec:semantic regularization}
In our pilot study~(Sec.~\ref{sec:preliminary exp}), the latent space complexity significantly increases as the tokenizer scales, which potentially leads to worse downstream AR generation for larger tokenizers. 
% \tianwei{The new motivation.} 
We hypothesize that larger tokenizers tend to capture excessive fine-grained low-level details for better reconstruction, resulting in overly complex latent token distributions, which makes it harder for AR models to learn the token dependencies effectively.

To address this, we introduce semantic
regularization to guide the tokenizer to encode a more semantically consistent latent space, which is less complex and easier for downstream generative modeling.
Specifically, we introduce a simple semantic regularization term alongside the tokenizer training objective.
The regularization aligns the intermediate features of the tokenizer decoder with the feature representations extracted from pre-trained frozen DINOv2~\cite{dinov2}.

Mathematically, let $f^{\text{dec}, l}$ be the output feature of the $l$-th layer of the Transformer decoder, $f^{\rm DINO}$ be the semantic features of a pretrained image encoder (here DINOv2-B~\cite{dinov2}). The semantic regularization can be represented as:
\begin{equation}
\label{eq:sem reg}
    \mathcal{L}_{\rm reg} = \frac{1}{N} \sum_{n=1}^N  {\rm sim} \Big( f^{\text{dec},l}_n, \phi(f^{\rm DINO}_n) \Big)
\end{equation}
where $N$ is the batch size, $n$ is the image index, $\text{sim}(\cdot, \cdot)$ is a cosine similarity function, and $\phi(\cdot)$ is an MLP that projects decoder feature $f^{\text{dec}, l}$ to match the channel dimension of $f^{\rm DINO}$. When training VQ tokenizers, we add the semantic regularization to the original VQGAN~\cite{vqgan, llamagen} objectives:

\begin{equation}\label{eq:loss}
    \mathcal{L}_{\text{total}} = \mathcal{L}_{\text{vqgan}} + \lambda \mathcal{L}_{\text{reg}},
\end{equation}
and we empirically set $\lambda=0.5$ in this work. 
Here $\mathcal{L}_{\text{vqgan}}$ is a combination of multiple losses
% :
, including $\mathcal{L}_{\text{recon}}$, 
the $l_2$ reconstruction loss on image pixels, $\mathcal{L}_{\text{percp}}$, the perceptual loss~\cite{LPIPS, johnson2016perceptual}, $\mathcal{L}_{\text{GAN}}$, PatchGAN~\cite{patchgan} adversarial loss, and $\mathcal{L}_{\text{VQ}}$~\cite{vqgan, vit_vqgan} the VQ codebook loss.

\begin{figure}[]
% \vspace{-0.1in}
    \centering
    \includegraphics[width=\linewidth]{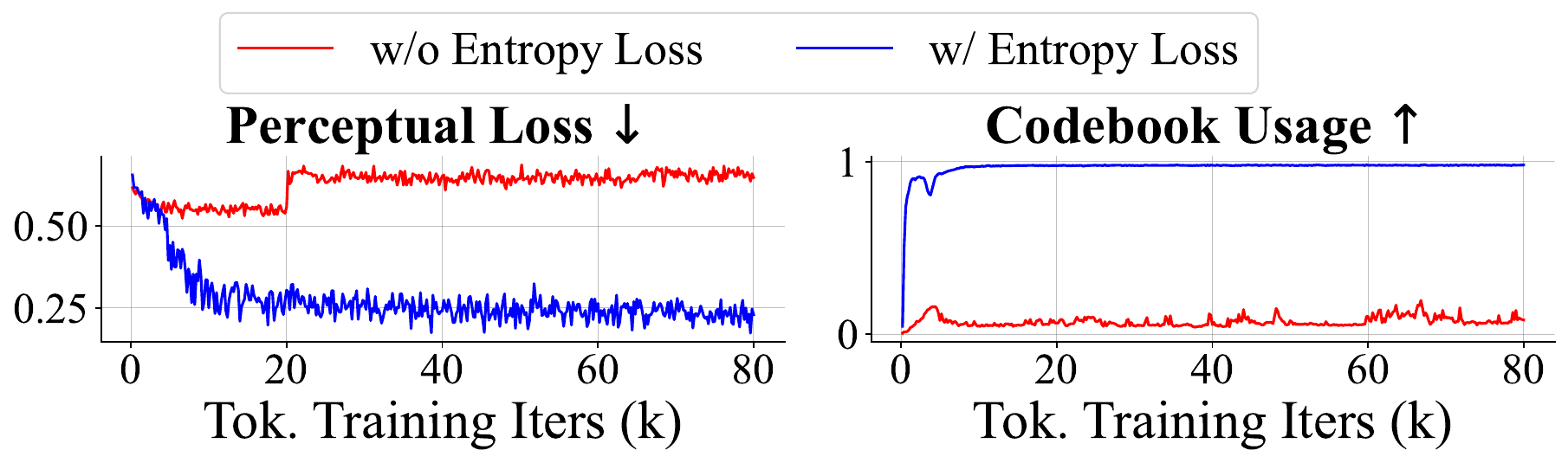}
    % \vspace{-0.30in}
    \caption{\textbf{Training curves for 2.9B XL-XXL tokenizers with and without entropy loss.} A 2.9B tokenizer does not converge without entropy loss. The entropy loss encourages high codebook usage and stabilizes training loss. 
    }
    \vspace{-5pt}
    \label{fig:billion_tok_entropy}
\end{figure}

\subsection{Entropy Loss for Billion-Level Tokenizers}
\label{sec:entropy loss}

When training a 2.9B tokenizer, we find that using the same training recipe as the 622M tokenizer leads to convergence failure for both perceptual loss and reconstruction loss,
and consistently low codebook usage. 
We hypothesize that low codebook usage accounts for the convergence difficulty. To address this, we incorporate entropy penalty~\cite{magvit-v2, yu2023magvit} to encourage higher codebook utilization: 
\begin{equation}
\mathcal{L}_{\text{entropy}} = \mathbb{E}_{\mathbf{z}}\left[H(\hat{\mathbf{z}}|\mathbf{z})\right] - H(\hat{\mathbf{z}})
\end{equation}
where $H(\cdot)$ denotes the Shannon entropy, $\mathbf{z}\in \mathbb{R}^{D}$ is the input for quantizer to be quantized to $\mathbf{\hat{z}}=\mathbf{c}_{i}\in \mathbb{R}^{D}$ and $\mathbf{c}_{i}$ is the $i$-th codebook vector. $\mathbb{E}_{\mathbf{z}}\left[H(\hat{\mathbf{z}}|\mathbf{z})\right]$ penalizes the uncertainty in quantization to reduce quantization error, and $- H(\hat{\mathbf{z}})$ encourages the codebook vectors to be selected more uniformly across the entire codebook. The detailed derivation can be found in our \textit{supp}. %Note that entropy penalty is \textbf{not} our contribution.
We find that the entropy penalty addresses the convergence difficulty of large tokenizers. As shown in Fig.~\ref{fig:billion_tok_entropy}, introducing entropy loss to the 2.9B tokenizer enables the codebook usage to quickly reach a high level, and the loss converges properly\footnote{We take perceptual loss as an example, and reconstruction loss shows a similar pattern}.

\begin{table}[ht]
\centering
\small
\begin{tabular}{l@{\hspace{1pt}}rrrrrr}
\toprule
Type & Enc./Dec. & Params. & Blocks & Heads & Dim. \\
\midrule
\multirow{5}{*}{1D Tok.} & S  & 26M  & 6   & 8   & 512   \\
 & B  & 115M  & 12   & 12   & 768   \\
 & L & 405M  & 24  & 16   & 1024     \\
 & XL  & 948M  & 36   & 20   & 1280   \\
 & XXL  & 1870M  & 48  & 24   & 1536   \\
\midrule
\multirow{3}{*}{2D Tok.} & S   & 19M  & 6   & 8   & 512   \\
& B & 86M & 12   & 12   & 768   \\
& L & 329M  & 24  & 16   & 1024   \\
% \midrule
% \multirow{4}{*}{AR Model} & B  & 111M & 12  & 12  & 768  \\
%  & L  & 343M & 24  & 16  & 1024  \\
%  & XL & 775M  & 36 & 20  & 1280   \\
%  & XXL  & 1.4B & 48  & 24  & 1536  \\
\bottomrule
\end{tabular}
\caption{\textbf{Architectures of the transformer variants for tokenizer encoder/decoder parts in our experiments.}  
We use Q-Former~\cite{blip2, DETR} for 1D tokenizers and ViT~\cite{vit} for 2D tokenizers.}
\label{tab:transformer setting}
% \vspace{-5pt}
\end{table}

\begin{figure}[]
% \vspace{-0.1in}
    \centering
    \includegraphics[width=\linewidth]{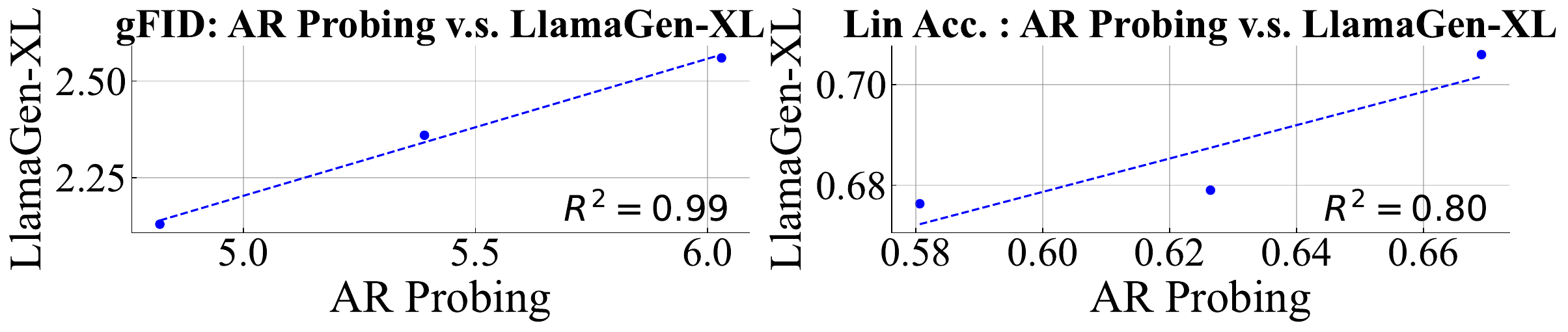}
    % \vspace{-0.30in}
    \caption{\textbf{Correlation between AR Probing Performance and Larger AR models.} For 3 tokenizers: S-S, S-L, and B-L, we present that as the tokenizer improves, the performance improvements of AR Probing correlate to the performance improvements of larger AR models. Therefore, the AR Probing can effectively indicate how the tokenizer affects downstream larger AR models with limited computational costs.
    }
    \vspace{-5pt}
    \label{fig:ar_probe_correlation}
\end{figure}

\section{Experiments}

\begin{figure*}[]
% \vspace{-0.1in}
    \centering
    \includegraphics[width=\linewidth]{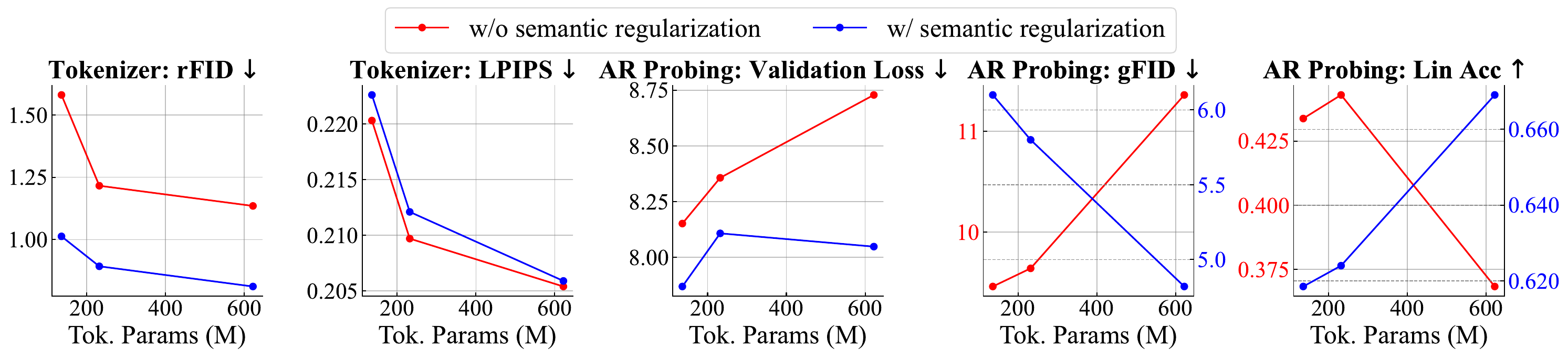}
    % \vspace{-0.30in}
   \caption{\textbf{Scaling trends of tokenizers for reconstruction, downstream generation and representation quality with and without semantic regularization.} By semantic regularization, \ours{} resolves the reconstruction \vs generation dilemma for tokenizer scaling in contrast to the vanilla version without semantic regularization. Moreover, \ours{} consistently improves the representation quality of downstream AR models by scaling up visual tokenizers. Note that in the last two figures, the red and blue curves correspond to different scales on the y-axis.
    }
    \vspace{-5pt}
    \label{fig:tok_scale_cmp}
\end{figure*}

\subsection{Settings}\label{subsec:settings}
For scaling up visual tokenizers, we follow the architecture configurations for the Transformers in \ours{} tokenizers as summarized in Tab.~\ref{tab:transformer setting}. 
We evaluate the tokenizers from three perspectives: reconstruction, downstream AR generation, and downstream AR representation quality. We use rFID and LPIPS~\cite{LPIPS} to evaluate reconstruction fidelity, gFID to evaluate generation performance, and linear probing to evaluate the representation quality of the downstream AR model. Our downstream AR models are LlamaGen~\cite{llamagen} with 1D absolute positional embedding. Our scaling experiments~(Sec.~\ref{sec:sem exp}) and ablation study~(Sec.~\ref{sec:arch exp}) use AR Probing~(111M AR model described in Sec.\ref{sec:AR Probe}) validation loss, gFID, and linear probing to reflect the learnability of tokens, generation performance, and representation quality, respectively. While in the system-level comparison (Sec.~\ref{sec:main cmp}), we train larger 1.4B AR models for comparison with previous work.
More details are in the supplementary material.

\noindent\textbf{Effectiveness of AR Probing.} As shown in Fig.~\ref{fig:ar_probe_correlation}, AR Probing performances including gFID and linear probing accuracy align with the larger LlamaGen-XL~\cite{llamagen} model results. Therefore, we use AR Probing throughout the following experiments except for the system-level comparison.

\subsection{Scaling with Semantic Regularization}
\label{sec:sem exp}

% \subsection{Experiment Setup}
\label{sec: setup scaling semantic}
We demonstrate that our proposed semantic regularization resolves the reconstruction \vs generation dilemma in scaling tokenizers.

\noindent\textbf{Model scaling with semantic regularization.} Results are shown in Fig.~\ref{fig:tok_scale_cmp}.~(1) Semantic regularization improves the reconstruction fidelity, indicated by lower rFID.~(2) More importantly, the AR Probing validation loss and gFID degrades for larger tokenizers without semantic regularization, showing the reconstruction \vs generation dilemma. The dilemma is addressed with semantic regularization, evidenced by the relatively constrained validation loss and consistently decreasing gFID.~(3) The Linear Probing results show that semantic regularization helps AR models to learn better representations as the tokenizer model scales up.

\begin{figure}[t]
% \vspace{-0.1in}
    \centering
    \includegraphics[width=\linewidth]{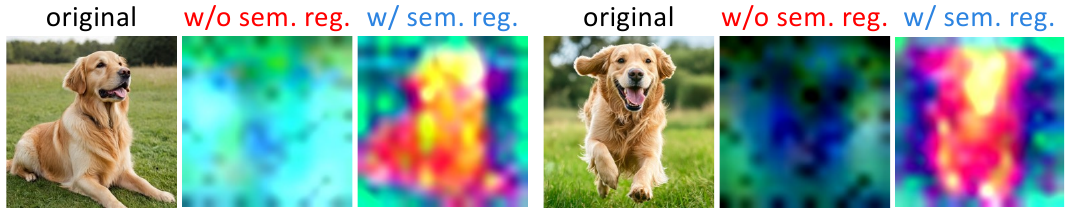}
    % \vspace{-0.30in}
    \caption{\textbf{Visualization 
    of tokenizer features with and without semantic regularization.} We compute PCA among the tokenizer features of a group of images of the same ``golden retriever'' class
    and visualize the first 3 PCA components.
    We observe that the latent space of vanilla tokenizers shows inconsistent features both within a single image or across multiple semantically similar images. In contrast, \ours{} encodes images with semantic consistency and thus reduces the latent space complexity for AR models.  
    }
    \vspace{-5pt}
    \label{fig:latent_vis}
\end{figure}
\noindent\textbf{Visualization for the tokenizer feature space.} We visualize the first 3 PCA components of the tokenizer features from the first Transformer decoder layer for a group of images. As shown in Fig.~\ref{fig:latent_vis}, we find the vanilla tokenizer encodes a latent space with limited semantic consistency, which potentially impairs its learnability for downstream AR models. In contrast, \ours{} presents semantically consistent patterns~(Fig.~\ref{fig:latent_vis}), indicating a meaningful and consistent latent space. 
\begin{table}[t]
\centering
\begin{tabular}{c|cc|cc}
\toprule
Enc./Dec. Size & rFID$\downarrow$ & LPIPS$\downarrow$ & gFID$\downarrow$ & Lin Acc.$\uparrow$  \\
\midrule
B-S & 0.98   & 0.221    & 6.56 & 64.5  \\
S-B & 0.94 & 0.214  & 5.65   & 59.8   \\
\midrule
S-L & 0.83   & 0.206     & 5.19  & 60.6   \\
B-L & 0.81    & 0.206     & 4.82  & 66.9   \\
\bottomrule
\end{tabular}
\caption{\textbf{The results for scaling encoder/decoder.} Prioritizing the scaling of decoders benefits downstream generation more than scaling encoders~(S-B v.s. B-S). But scaling encoders can still bring significant improvements~(S-L v.s. B-L).}
\label{tab:cmp_enc_dec}
\vspace{-5pt}
\end{table}

% \begin{table}[h]
% \small
% \begin{tabular}{c|cc|ccc}
% \toprule
% Enc./Dec.& rFID$\downarrow$ & LPIPS$\downarrow$ & gFID$\downarrow$ & Lin Acc.$\uparrow$ & $\mathcal{L}_{\text{ce}}\downarrow$ \\
% \midrule
% B-S & 0.98   & 0.221    & 6.56 & 64.5  & 7.98   \\
% S-B & 0.94 & 0.214  & 5.65   & 59.8   & 7.80  \\
% \midrule
% S-L & 0.83   & 0.206     & 5.19  & 60.6   & 8.03  \\
% B-L & 0.81    & 0.206     & 4.82  & 66.9   & 8.05   \\
% \bottomrule
% \end{tabular}
% \caption{\textbf{The results for scaling encoder/decoder.} Prioritizing the scaling of decoders benefits downstream generation more than scaling encoders~(S-B v.s. B-S). But scaling encoders can still bring significant improvements~(S-L v.s. B-L).}
% \label{tab:cmp_enc_dec}
% \end{table}

\begin{figure}[h]
% \vspace{-0.1in}
    \centering
    \includegraphics[width=\linewidth]{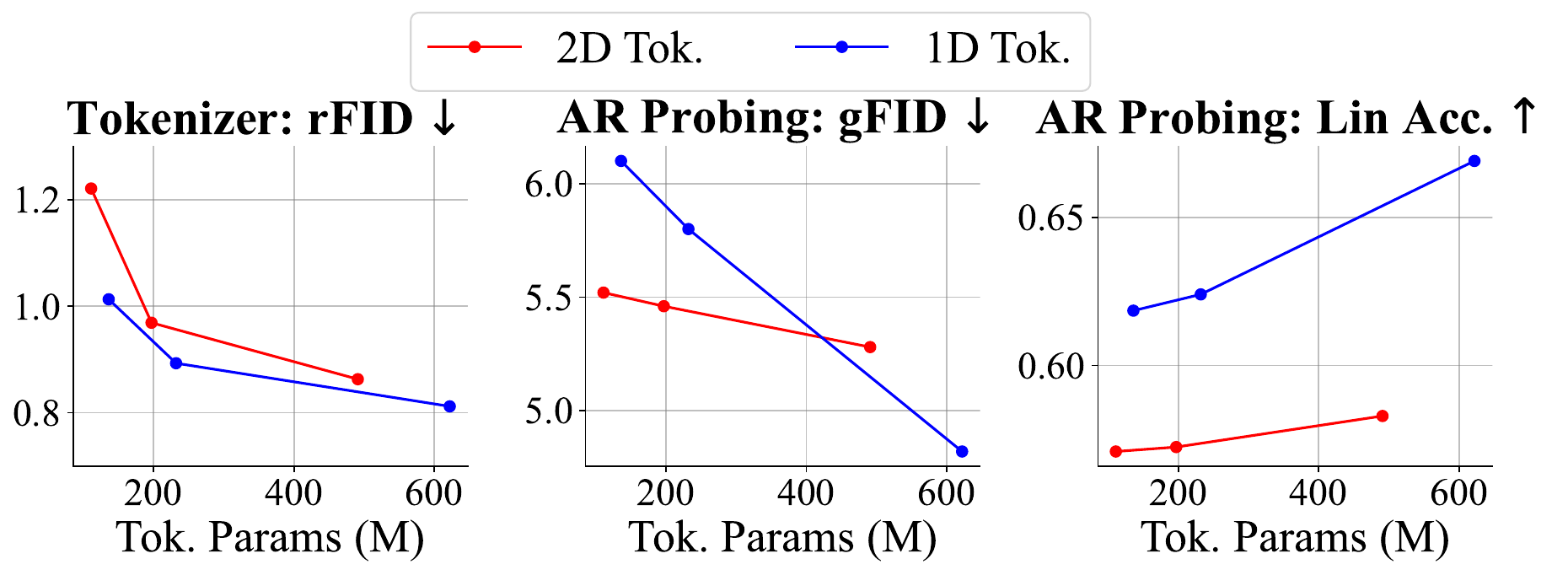}
    % \vspace{-0.30in}
    \caption{
    \textbf{Scalability comparison for 1D and 2D tokenizers.} Using the same training setting, 1D tokenizers shows better reconstruction~(rFID) and downstream representation quality~(AR Probing: Lin Acc.). For downstream generation~(gFID), 1D tokenizers present a steeper improving trend than 2D tokenizers.
    }
    \vspace{-5pt}
    \label{fig:tok_size_scale_2d_1d}
\end{figure}

\begin{table*}
\centering
% \fontsize{9.8pt}{11pt}\selectfont
\fontsize{9pt}{11pt}\selectfont
\begin{tabular}{l@{\hspace{4pt}}l@{\hspace{4pt}}rcl|l@{\hspace{3pt}}rrl@{\hspace{4pt}}c}
\toprule
Tokenizer & 
\multicolumn{2}{c}{Tok. Type/Param.} 
% Tok.Type. &
% Param.
& \#Tokens & rFID$\downarrow$ & \multicolumn{2}{c}{Generator  Model/Param.} & Type & gFID$\downarrow$ & Acc.$\uparrow$ \\
\midrule
\multicolumn{10}{c}{\textit{\textbf{Continuous token modeling}}} \\
\midrule

VAE~\cite{ldm_stable_diffusion} & KL$^{\dag}$ & 55M & 4096 & 0.27 & LDM-4~\cite{ldm_stable_diffusion}     & 400M & Diff. & 3.60 & - \\
\multirow{3}{*}{SD-VAE~\cite{sd-vae-ft-ema}} & \multirow{3}{*}{KL$^{\dag}$} & \multirow{3}{*}{84M} & \multirow{3}{*}{1024} & \multirow{3}{*}{0.62} & DiT-XL/2~\cite{dit}    & 675M & Diff. & 2.27 & - \\
&  &  &  &  & SiT-XL/2~\cite{ma2024sit}    & 675M  & Diff. & 2.06 & - \\
&  &  &  &  & SiT-XL/2 + REPA~\cite{repa}    & 675M  & Diff. & 1.42 & 74.6 \\
% data source: VA-VAE
VA-VAE~\cite{lightningDiT} & KL & 70M & 256 & 0.28 & LightningDiT~\cite{lightningDiT}    & 675M & Diff. & 1.35 & - \\
%\midrule
VAE~\cite{MAR} & KL & 66M & 256 & 0.53 & MAR-H~\cite{MAR}  & 943M & AR+Diff. & 1.55 & 60.0$^{\diamond}$ \\
\midrule
\multicolumn{10}{c}{\textit{\textbf{Discrete token modeling}}} \\
\midrule
VQGAN~\cite{chang2022maskgit} & VQ & 66M & 256 & 2.28 & MaskGIT~\cite{chang2022maskgit}     & 227M & Mask. &  6.18$^{\star}$ & - \\
% data source: TiTok, the model size is measured
TiTok-S~\cite{titok} & VQ & 72M  & 128 & 1.71 & MaskGIT-UViT-L ~\cite{chang2022maskgit, UViT}  & 287M & Mask. & 1.97 & - \\
TiTok-L~\cite{titok} & VQ & 641M  & 32 & 2.21 & MaskGIT-ViT ~\cite{chang2022maskgit}  & 177M & Mask. & 2.77 & - \\
% data source: BiGR, the model size is measured. The rFID is from the openreview rebuttal results.
\multirow{2}{*}{B-AE-d32~\cite{bigr}} & \multirow{2}{*}{LFQ} & \multirow{2}{*}{66M} & \multirow{2}{*}{256} & \multirow{2}{*}{1.69} & BiGR-XXL-d32~\cite{bigr}  & 1.5B & AR+Diff & 2.36 & - \\
 &  &  &  &  & BiGR-XL-d32~\cite{bigr}  & 799M & AR+Diff & - & 69.8 \\
\midrule
\multirow{2}{*}{VAR-Tok.~\cite{var}} & \multirow{2}{*}{MSRQ$^\dag$}  & \multirow{2}{*}{109M} & \multirow{2}{*}{680} & \multirow{2}{*}{1.00$^{\ddag}$} & VAR-$d24$~\cite{var}       & 1.0B & VAR & 2.09 & - \\
&&&&& VAR-$d30$~\cite{var}       & 2.0B & VAR & 1.92 & - \\
ImageFolder~\cite{imagefolder} & MSRQ & 176M & 286 & 0.80$^{\ddag}$ & ImageFolder-VAR~\cite{imagefolder}    & 362M & VAR & 2.60 & - \\
\midrule
% data source: Correct rFID is from the original VQGAN repo.
VQGAN~\cite{vqgan} & VQ & 23M & 256 & 4.98 & Taming-Tran.~\cite{vqgan}       & 1.4B & AR & 15.78$^{\star}$ & - \\
ViT-VQGAN~\cite{vit_vqgan} & VQ & 64M & 1024 & 1.28 & VIM-Large~\cite{vit_vqgan}& 1.7B & AR & 4.17$^{\star}$ & - \\
RQ-VAE~\cite{rq} & RQ & 66M & 256 & 3.20 & RQTran.~\cite{rq}        & 3.8B & AR & 7.55$^{\star}$ & - \\
% data source: OpenMAGVIT-v2, param is measured
Open-MAGVIT2~\cite{openmagvit-v2} & LFQ & 133M & 256 & 1.17 & Open-MAGVIT2-XL~\cite{openmagvit-v2}        & 1.5B & AR & 2.53 & - \\
% data source: IBQ, param is measured
IBQ~\cite{ibq_scale} & IBQ & 128M & 256 & 1.37 & IBQ-XXL~\cite{ibq_scale}        & 2.1B & AR & 2.05 & - \\
% data source: LlamaGen and BiGR for acc
\multirow{2}{*}{LlamaGen-Tok.~\cite{llamagen}} & \multirow{2}{*}{VQ} & \multirow{2}{*}{72M} & \multirow{2}{*}{256} & \multirow{2}{*}{2.19} & LlamaGen-L~\cite{llamagen}        & 343M & AR & 3.81 & 40.5$^{\diamond}$ \\
 & & & &  & LlamaGen-XXL~\cite{llamagen}        & 1.4B & AR & 3.09 & - \\
LlamaGen-Tok.~\cite{llamagen} & VQ & 72M & 576 & 0.94 & LlamaGen-XXL~\cite{llamagen}        & 1.4B & AR & 2.34 & - \\
\midrule
\ours{}-B-L & VQ & 622M & 256 & 0.51$^{\ddag}$ & LlamaGen-B~(1d)~\cite{llamagen}    & 111M & AR & 3.33 & 67.7 \\
\ours{}-S-S & VQ & 136M & 256 & 1.01 & LlamaGen-B~(1d)~\cite{llamagen}    & 111M & AR & 4.05  & 62.6  \\
\ours{}-S-B & VQ & 232M & 256 & 0.89 & LlamaGen-B~(1d)~\cite{llamagen}    & 111M & AR & 3.83  & 62.9  \\
\multirow{2}{*}{\ours{}-B-L} & \multirow{2}{*}{VQ} & \multirow{2}{*}{622M} & \multirow{2}{*}{256} & \multirow{2}{*}{0.81} & LlamaGen-B~(1d)~\cite{llamagen}    & 111M & AR & 3.26  & 67.6 \\
 & & & &  & LlamaGen-XXL~(1d)~\cite{llamagen}    & 1.4B & AR & 2.03$^{\star}$ & 69.4 \\
\multirow{2}{*}{\ours{}{-XL-XXL}} & \multirow{2}{*}{VQ} & \multirow{2}{*}{2.9B} & \multirow{2}{*}{256} & \multirow{2}{*}{0.79} & LlamaGen-B~(1d)~\cite{llamagen}    & 111M & AR & 3.15 & 72.0 \\
& & & &  &  LlamaGen-XXL~(1d)~\cite{llamagen}    & 1.4B & AR & 1.98$^{\star}$ & 74.0 \\
\bottomrule
\end{tabular}
\caption{\textbf{System-level comparison for tokenizers and downstream generation models on ImageNet 256$\times$256.} For gFID, we present the lowest value between w/ or w/o CFG scenarios. \dag: Training set includes data besides ImageNet. \ddag: Using frozen DINO~\cite{dino} for discriminator, which largely improves rFID. $\star$: Without classifier-free-guidance. $\diamond$: Data from BiGR~\cite{bigr}. 
}
\vspace{-5pt}
\label{tab:main_cmp}
\end{table*}

\subsection{Asymmetric 1D Tokenizer is More Scalable}
\label{sec:arch exp}

\noindent\textbf{Tokenizer decoder deserves more parameters.} To determine whether the decoder or encoder should be prioritized when scaling up, we compare S-B\footnote{X-Y tokenizer denotes X-sized encoder and Y-sized decoder. For example, S-B indicates Small encoder-Base decoder structure} and B-S tokenizers in Tab.~\ref{tab:cmp_enc_dec}, both trained under the same setting for 100 epochs. Our results show that scaling decoders, rather than encoders, leads to greater improvements in both reconstruction and downstream generation, suggesting that decoder scaling should be prioritized.

\noindent\textbf{Scaling tokenizer encoder is also important.} While prioritizing the scaling of tokenizer decoders yields significant benefits, we also find that scaling tokenizer encoders can further enhance downstream models. In Tab.~\ref{tab:cmp_enc_dec}, we show that a B-L tokenizer gains significant improvements compared to an S-L tokenizer.
Therefore, we recommend scaling both encoders and decoders while maintaining a larger decoder than the encoder for optimal performance.

\noindent\textbf{1D tokenizers are more scalable than 2D tokenizers.} We train S-S, S-B and B-L 1D/2D tokenizers with the same setting with semantic regularization. As shown in Fig.~\ref{fig:tok_size_scale_2d_1d}, 1D tokenizers consistently achieve better rFID and AR Probing linear probing accuracy than 2D tokenizers. For AR Probing gFID, the 1D tokenizers exhibit a steeper scaling trend, eventually surpassing 2D tokenizers as the model scales. We attribute the superior scalability of 1D tokenizers to the reduced inductive bias.

\subsection{System-level Comparison}
\label{sec:main cmp}

\noindent\textbf{Experiment Settings.}
Using \ours{} for tokenization, we scale the training of LlamaGen~\cite{llamagen} AR models on $256\times 256$ ImageNet training set for 300 epochs to compare with other methods. We do not use AdaLN~\cite{dit, var} as it is specific for class-conditional generation. We provide the results of a B-L tokenizer trained with DINO discriminator~\cite{var, imagefolder} to fairly compare rFID. But in practice we find DINO discriminator provides limited improvement for LPIPS and may affect the training stability of billion-scale tokenizers. Therefore, we exclude it from our main design.

\noindent\textbf{Results.}
As shown in Tab.~\ref{tab:main_cmp}, our 2.9B \ours~achieves state-of-the-art reconstruction performance (rIFD) among all discrete tokenizers. Furthermore, with our 2.9B tokenizer, the downstream 1.4B AR model achieves state-of-the-art image generation performance (gFID) among LLM-style autoregressive next-token-prediction models. VAR~\cite{var} predicts images with next-scale prediction rather than next-token-prediction, which is less compatible with language models. Our model achieves comparable gFID to VAR~\cite{var} with a simple LLM-style downstream AR generator without incorporating vision-specific designs like VAR. Moreover, this 1.4B AR model trained on the 2.9B tokenizer achieves state-of-the-art linear probing accuracy via visual generative pretraining\footnote{REPA~\cite{repa} achieves better representation by directly distilling pretrained representations to the generation model, which is not a fair comparison with ours as we do not leverage the supervision for AR training.}. This indicates that our \ours{} helps the downstream generation model to learn better representations. The high-quality representation learned from generative pre-training may also help unify generation and understanding for future native multimodal models.

\subsection{Discussion and Ablation Study}
\label{sec:ablation}

\begin{table}[t]
\centering
\begin{tabular}{crcc}
\toprule
Decoder\textbackslash AR Model Size & B & L & XXL \\
\midrule
B & 3.7\% & 2.3\% & 1.3\% \\
L & 11.2\% & 7.0\% & 3.4\% \\
XXL & 32.4\% & 20.3\% & 9.9\% \\
\bottomrule
\end{tabular}
\caption{\textbf{Ratio of time consumptions for tokenizer decoding during image generation.} When we use a 2.9B XLXXL tokenizer for a 1.4B LlamaGen-XXL AR model, the tokenizer decoding only takes 9.9\% of the total inference time. 
}
\label{tab:time cnt}
\end{table}

\begin{table}[ht]
\centering
\begin{tabular}{c|cc|cc}
\toprule
Align. Layer $l$ & rFID$\downarrow$ & LPIPS$\downarrow$ & gFID$\downarrow$ & Lin Acc.$\uparrow$ \\
\midrule
2 & 1.06 & 0.224   & 6.26 & 63.4  \\
3 & 1.01 & 0.223  & 6.10 & 61.9 \\
4 & 1.07 & 0.223  & 6.07 & 58.6 \\
\bottomrule
\end{tabular}
\caption{\textbf{Layer $l$ for semantic regularization (S-S tokenizer).} Smaller $l$ brings better downstream AR model representations but can sacrifice reconstruction and downstream generation quality. We choose $l$=3 by default for more balanced performance.
}

\label{tab:layer index}
\end{table}

\begin{table}[ht]
\begin{tabular}{l|cc|cc}
\toprule
Sem. Enc. & rFID$\downarrow$ & LPIPS$\downarrow$ & gFID$\downarrow$ & Lin Acc.$\uparrow$ \\
\midrule
CLIP~\cite{CLIP, clipdfn} & 0.91 & 0.210  & 6.35 & 61.4  \\
SigLIP~\cite{siglip} & 0.92 & 0.210  & 6.20 & 56.7 \\
DINOv2-B~\cite{dinov2} & 0.85 & 0.212  & 5.55 & 64.4 \\
\bottomrule
\end{tabular}
\caption{\textbf{Ablation study for the choice of pretrained semantic encoders (S-B tokenizer).} DINOv2-B delivers the best performance among all models.}
\label{tab:sem enc}

\end{table}
\begin{table}[ht]
\centering
\begin{tabular}{c|cc|cc}
\toprule
Sem. Reg. $\lambda$ & rFID$\downarrow$ & LPIPS$\downarrow$ & gFID$\downarrow$ & Lin Acc.$\uparrow$ \\
\midrule
0.25 & 1.28 & 0.226   & 6.27 & 57.0  \\
0.50 & 1.22 & 0.228  & 6.39 & 58.6 \\
0.75 & 1.27 & 0.236  & 6.29 & 58.6 \\
1.00 & 1.38 & 0.239  & 6.27 & 62.5 \\
\bottomrule
\end{tabular}
\caption{\textbf{Ablation Study for the semantic regularization weight (S-S tokenizer).} A strong semantic regularization weight leads to worse reconstruction but better downstream representation. We choose $\lambda=0.5$ by default for more balanced performance.}
\label{tab:sem weight}

\end{table}

\smallskip \noindent \textbf{Discussion on generation costs.}
When generating an image, AR models take multiple passes to predict tokens, while tokenizers only need one forward pass. Therefore, the time consumption for decoding tokens to images is relatively small compared to AR models. We record the ratio of time spent on tokenizer decoding for different tokenizer/AR models in Tab.~\ref{tab:time cnt}. For a 1.4B AR model, our largest 2.9B tokenizer takes only $\sim$10\% of the total inference time.

\smallskip
\noindent\textbf{Searching the best layer for semantic regularization.} We search $l$, the layer's index in the Transformer decoder before intermediate features are extracted to calculate semantic regularization in Eq.~\ref{eq:sem reg}. As shown in Tab.~\ref{tab:layer index}, varying $l$ presents a trade-off between gFID and the Lin Acc. for AR Probing. Smaller $l$ means stricter regularization for the latent space so that the downstream generation models learn better representation. However, smaller $l$ also sacrifices generation quality. We choose $l=3$ for a more balanced rFID, gFID, and linear probing accuracy for all tokenizers.
% , because tuning $l$ is costly and the results are less predictable or generalizable. 

\smallskip
\noindent\textbf{Exploring pretrained semantic encoder choices.} We compare CLIP~(DFN)~\cite{clipdfn, CLIP}, SigLIP-400M~\cite{siglip} and DINOv2-B~\cite{dinov2} as the source of semantic regularization for S-B tokenizers. As shown in Tab.~\ref{tab:sem enc}, utilizing DINOv2-B as the semantic encoder for regularization produces the best tokenizer for reconstruction, downstream class conditional generation and representation quality.

\smallskip
\noindent\textbf{Exploring weights for semantic regularization.} 
We study the effects of different regularization weights $\lambda$ (Eq.~\ref{eq:loss}), from $0.25$ to $1.00$. As shown in Tab.~\ref{tab:sem weight}, a large $\lambda$~($0.75, 1.00$) will damage the reconstruction quality but benefits the linear probing accuracy, whereas smaller $\lambda$~($0.25$) results in suboptimal rFID and linear probing accuracy. We choose the more balanced $\lambda=0.5$ as a default for all tokenizers.

\section{Conclusion}
In this work, we study and address the reconstruction \vs generation dilemma for scaling visual tokenizers. We identify that the dilemma stems from increasing latent space complexity in larger tokenizers. 
We propose semantic regularization to effectively regularize the tokenizer latent space by injecting pre-trained representations to align with tokenizer features in training.
The semantic regularization, together with several key practices we explored, lead to the first 3B tokenizer, \ours{}, that achieves state-of-the-art reconstruction, downstream AR generation, and downstream AR representation quality. Please refer to discussions on limitations and future work in supplementary materials.

\clearpage
\section*{Acknowledgments}
This work is partially supported by the National Nature Science Foundation of China (No. 62402406). 

The authors also sincerely thank Qihang Yu and Liang-Chieh Chen for their valuable discussions during the development of GigaTok. 
% \clearpage

{
    \small
    \bibliographystyle{ieeenat_fullname}
    \bibliography{main}
}

\clearpage
\appendix

\clearpage
\setcounter{page}{1}
\maketitlesupplementary

\section{Limitations and Future Work}

This study primarily focuses on scaling tokenizers for class-conditional image generation. While we have demonstrated the effectiveness of \ours{} for downstream class-conditional generation, expanding the scope to include text-conditional image generation or video generation remains an open avenue for future work. Additionally, unlike CNN-based 2D tokenizers, 1D Transformer-based tokenizers are not directly applicable to multiple resolutions without additional training adjustments. This challenge presents an important direction for further exploration. Besides scaling the model sizes of tokenizers, the effect of scaling training data, codebook dimension and codebook size for downstream autoregressive generation are left for future research.

\section{Configurations for AR models}

\begin{table}[h]
\centering
\small
\begin{tabular}{@{\hspace{1pt}}rrrrrr}
\toprule
Size & Params. & Blocks & Heads & Dim. \\
\midrule
B  & 111M & 12  & 12  & 768  \\
L  & 343M & 24  & 16  & 1024  \\
XL & 775M  & 36 & 20  & 1280   \\
XXL  & 1.4B & 48  & 24  & 1536  \\
\bottomrule
\end{tabular}
\caption{\textbf{Architectures of the LLamaGen models in our experiments.}}
\label{tab:llama setting}
\end{table}

\noindent\textbf{AR model training.} We scale up the training of downstream  Llama-style~\cite{llama, llamagen} AR models to compare generation performance with other models.
For model training, we use WSD learning rate scheduler~\cite{hu2024minicpm, wsd_scale} with $1\times10^{-4}$ base learning rate, $0.2$ decay ratio and 1 epoch warm-up. We do not use AdaLN~\cite{dit, var} as it is specific for class-conditional generation. We use a batch size of 256 for training the B, L and XL models and a 512 batch size for training the XXL model. Our AR models are trained for 300 epochs on the $256\times 256$ ImageNet training set.

\noindent\textbf{CFG for gFID.}
Since gFID of GPT models can be largely affected by classifier free guidance~(CFG)~\cite{llamagen, ldm_stable_diffusion} and often has an optimal CFG~\cite{llamagen}, for fair comparison, we search the optimal CFG using zero-order search with a step of 0.25 and use the lowest gFID as the final value. For AR Probing, we use constant CFG scheduling for simplicity. For system-level comparison, we use a step function for CFG scheduling inspired by \cite{cfg_interval}. Specifically, the AR models predict the first 18\% tokens without CFG, \ie, $\text{CFG}=1$ for better diversity, and use CFG for the remaining tokens for better visual quality. Interestingly, we find that the 1.4B LlamaGen model achieves the best gFID without CFG.

\section{Detailed \ours{} Implementation}

Please refer to Tab.~\ref{tab:training_setting} for training details.

\begin{figure}[t]
% \vspace{-0.1in}
    \centering
    \includegraphics[width=\linewidth]{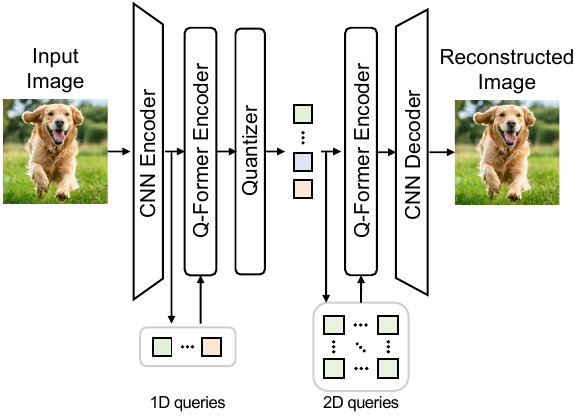}
    % \vspace{-0.30in}
    \caption{\textbf{The architecture of \ours{} with Q-Former.}
    }
    % \vspace{-0.3in}
    \label{fig:1d_qformer}
\end{figure}

\begin{table*}[ht]
\centering
\begin{tabular}{l|ccccc}
\toprule
\textbf{Configuration} & \textbf{S-S} & \textbf{S-B} & \textbf{S-L} & \textbf{B-L} & \textbf{XL-XXL} \\
\midrule
Q-Former Encoder depth           & 6    & 6    & 6    & 12    & 36    \\
Q-Former Encoder heads           & 8    & 8    & 8    & 12    & 20    \\
Q-Former Encoder dim.             & 512   & 512  & 512  & 768  & 1280  \\
Q-Former Decoder depth           & 6    & 12    & 24    & 24    & 48    \\
Q-Former Decoder heads.             & 8   & 12  & 16  & 16  & 24  \\
Q-Former Decoder dim.            & 512   & 768  & 1024  & 1024  & 1536  \\
Params~(M)            & 136   & 232  & 533  & 622  & 2896  \\
\midrule
Codebook size &\multicolumn{5}{c}{16384}\\
Codebook dimension &\multicolumn{5}{c}{8}\\
\#Tokens &\multicolumn{5}{c}{256}\\
\midrule
Training epochs & 100 & 200 & 200 & 200 & 300 \\
Batch size & 128 & 128 & 256 & 256 & 256 \\
Alignment Layer $l$ & \multicolumn{5}{c}{3} \\
Learning rate schedule & \multicolumn{5}{c}{Cosine Decay} \\
Base learning rate & \multicolumn{5}{c}{$1\times10^{-4}$} \\
Minimum learning rate &\multicolumn{5}{c}{$1\times 10^{-5}$}\\
LR warm-up iterations & 0 & 0 & 0 & 0 & 5000 \\
Optimizer & \multicolumn{5}{c}{AdamW\cite{adamw}} \\
Opt. momentum & \multicolumn{5}{c}{$\beta_1 = 0.9, \beta_2=0.95$} \\
Entropy Loss weight& 
0 & 0 & 0 & 0 & $5\times 10^{-3}$ \\

\bottomrule
\end{tabular}
\caption{\ours{} configuration and default training details}
\label{tab:training_setting}
\end{table*}

\begin{figure}[]
% \vspace{-0.1in}
    \centering
    \includegraphics[width=\linewidth]{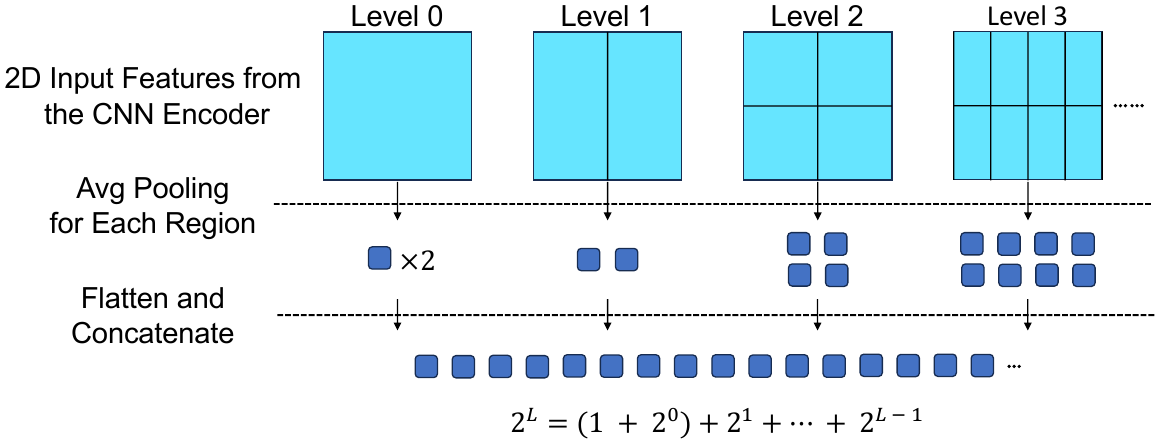}
    % \vspace{-0.30in}
    \caption{\textbf{Initialization of 1D queries in Q-Former modules.}
    }
    % \vspace{-0.3in}
    \label{fig:1d_query_init}
\end{figure}

\noindent\textbf{Q-Fomrer in \ours{}.} \ours{} utilizes Q-Former~\cite{blip2, DETR} to build 1D tokenizers, as shown in Fig.~\ref{fig:1d_qformer}. For Q-Former encoder in \ours{}, we initialize the 1D queries initialized from the 2D input features of the CNN encoder using a multi-level average pooling strategy, as shown in Fig.~\ref{fig:1d_query_init}. Specifically, for the same 2D input features, we spatially divide them with different granularity at different levels, and perform average pooling for every divided region at each level. The pooled features are flattened and concatenated from level 0 to the last level. Therefore, a 1D token sequence with $2^L$ length can be initialized with $L$ levels from 2D input features. At the decoding stage, the 2D queries are all initialized from the first 1D latent feature.

\noindent\textbf{Entropy Loss for VQ Tokenizers.} While entropy loss~\cite{magvit-v2, yu2023magvit} is discussed for LFQ~\cite{magvit-v2}, its application to VQ tokenizers is less commonly explained. We provide a detailed derivation of the entropy loss specifically for VQ tokenizers.
Mathematically, for quantization process from continuous vector $\mathbf{z}\in \mathbb{R}^{D}$ to quantized vector $\mathbf{\hat{z}}=\mathbf{c}_{i}\in \mathbb{R}^{D}$ where $\mathbf{c}_{i}$ is the $i$-th codebook vector from codebook $\mathbf{C}\in \mathbb{R}^{N\times D}$, we assume this process is statistical and follows the following distribution: 
\begin{equation}
    p(\hat{\mathbf{z}}= \mathbf{c}_{i}|\mathbf{z}) \triangleq \text{softmax}(-l_{2}(\mathbf{z}, \mathbf{C}))[i]
\end{equation}
where $l_2(\mathbf{z}, \mathbf{C})\in \mathbb{R}^{N}$ is the $L_2$ distance between $\mathbf{z}$ and all the codebook vectors. Then, minimization of the quantization error can be partially achieved by minimizing the expectation of entropy $\mathbb{E}_{\mathbf{z}}\left[H(\hat{\mathbf{z}}|\mathbf{z})\right]$, which can be understood as maximizing the prediction confidence for $p(\hat{\mathbf{z}}|\mathbf{z})$. To encourage higher codebook utilization, we aim to make the average appearance probability of codebook vectors more uniform. This is achieved by maximizing the entropy $H(\hat{\mathbf{z}})$, Therefore, the optimization of the two entropy terms leads to the final entropy loss equation:
\begin{equation}
    \mathcal{L}_{\text{entropy}} = \mathbb{E}_{\mathbf{z}}\left[H(\hat{\mathbf{z}}|\mathbf{z})\right] - H(\hat{\mathbf{z}})
\end{equation}
In practice, to calculate $H(\hat{\mathbf{z}})$, we estimate $p(\hat{\mathbf{z}}= \mathbf{c}_{i})$ by $p(\hat{\mathbf{z}}= \mathbf{c}_{i}) = \mathbb{E}_{\mathbf{z}}\left[p(\hat{\mathbf{z}}= \mathbf{c}_{i}|\mathbf{z})\right]$.
Note that entropy loss is \textbf{not} our contribution. We only provide a detailed definition of entropy loss in VQ scenarios for better understanding.

\noindent\textbf{Additional implementation details.}
To stabilize the training of our tokenizer with a hybrid architecture, we initially use a shortcut feature reconstruction trick at the first 15k iterations of the tokenizer training. But we later found that this trick can be replaced with a simple 
1-epoch learning rate warmup combined with entropy loss~\cite{magvit-v2, vqgan}. Specifically for this trick, we additionally give the output feature of the CNN encoder to the CNN decoder directly to be trained for reconstruction, and also align the output feature of the Transformer decoder to the output feature of the CNN encoder, besides the original training objectives. Note that this strategy is complex and can even hinder performance for XL-XXL tokenizers. We recommend using the learning rate warmup combined with entropy loss~\cite{magvit-v2, vqgan} instead, for both XL-XXL tokenizer and the smaller ones. Additionally, we utilize the rotation trick~\cite{rotation_trick} for all tokenizers, though we observe its effect on performance to be limited for our tokenizer.
The implementation of the semantic regularization is partially inspired by REPA~\cite{repa}.

\begin{table*}
\centering
\fontsize{9pt}{11pt}\selectfont
\begin{tabular}{l@{\hspace{4pt}}r@{\hspace{4pt}}l@{\hspace{4pt}}c@{\hspace{3pt}}c@{\hspace{4pt}}c@{\hspace{4pt}}|c@{\hspace{4pt}}c@{\hspace{4pt}}l@{\hspace{4pt}}c@{\hspace{4pt}}c@{\hspace{4pt}}c@{\hspace{4pt}}c}
\toprule
Tokenizer & Param. & rFID$\downarrow$ & LPIPS$\downarrow$ & PSNR$\uparrow$ & SSIM$\uparrow$ & AR Model & Param. & gFID$\downarrow$ & Acc.$\uparrow$ & IS$\uparrow$ & Precision$\uparrow$ & Recall$\uparrow$ \\
\midrule
LlamaGen-Tok.~\cite{llamagen} & 72M & 2.19 & -  & 20.79  & 0.675  & LlamaGen-B~\cite{llamagen} & 111M & 5.46 & - & 193.61   & 0.83   & 0.45 \\
\midrule
\ours{}-S-S & 136M & 1.01 & 0.2226  & 20.74  & 0.670  & LlamaGen-B~(1d)~\cite{llamagen} & 111M & 4.05 & 62.6 & 240.61   & 0.81   & 0.51  \\
\ours{}-S-B & 232M & 0.89 & 0.2121  & 20.93  & 0.677  & LlamaGen-B~(1d)~\cite{llamagen} & 111M & 3.83 & 62.9 & 233.31   & 0.83  & 0.51 \\
\multirow{2}{*}{\ours{}-B-L} & \multirow{2}{*}{622M} & \multirow{2}{*}{0.81} & \multirow{2}{*}{0.2059}  & \multirow{2}{*}{21.21}  & \multirow{2}{*}{0.685}  & LlamaGen-B~(1d)~\cite{llamagen} & 111M & 3.26 & 67.6 & 221.02  & 0.81  & 0.56  \\
 & & &  &  &  & LlamaGen-XXL~(1d)~\cite{llamagen} & 1.4B & 2.03$^{\star}$ & 69.4 & 238.52  & 0.80  & 0.63  \\
 \ours{}-B-L & 622M & 0.51$^{\ddag}$ & 0.206  & 21.32  & 0.691  & LlamaGen-B~(1d)~\cite{llamagen} & 111M & 3.33 & 67.7 & 265.43  &  0.80  & 0.56  \\
\multirow{2}{*}{\ours{}{-XL-XXL}} & \multirow{2}{*}{2.9B} & \multirow{2}{*}{0.79} & \multirow{2}{*}{0.1947}  & \multirow{2}{*}{21.65}  & \multirow{2}{*}{0.699}  & LlamaGen-B~(1d)~\cite{llamagen} & 111M & 3.15 & 72.0 & 224.28  & 0.82  & 0.55  \\
& & & & & & LlamaGen-XXL~(1d)~\cite{llamagen} & 1.4B & 1.98$^{\star}$ & 74.0 & 256.76  & 0.81  & 0.62  \\
\bottomrule
\end{tabular}
\caption{\textbf{Full results for our tokenizers and AR models on ImageNet 256$\times$256.} For gFID, we present the lowest value between w/ or w/o CFG scenarios. \ddag: Using frozen DINO~\cite{dino} for discriminator, which largely improves rFID. $\star$: Without classifier-free-guidance.
}
\label{tab:full results}
\end{table*}

\section{Full Evaluation Results and Analysis}

Here we present the full evaluation results for the tokenizers and downstream AR models, as summarized in Tab.~\ref{tab:full results}. We observe that scaling up visual tokenizers consistently improves the reconstruction quality across multiple metrics. Interestingly, for the 1.4B AR model, the lowest gFID is obtained without applying any CFG. This phenomenon is also observed in the concurrent work FlexTok~\cite{flextok}, despite significant differences between \ours{} and FlexTok. We hypothesize that semantic regularization might be the primary contributing factor for this phenomenon.

\noindent\textbf{Discussion on Scaling and Enhancing the Discriminator.} Recently, VAR~\cite{var}, ImageFolder~\cite{imagefolder}, and the concurrent work UniTok~\cite{ma2025unitok} have begun leveraging DINO-based discriminators~\cite{dino,dinov2} to enhance tokenizer training, achieving impressive improvements in rFID scores. We have also experimented with the same DINO discriminator configuration as VAR. Our results indicate that although rFID scores improve, the downstream generation quality improvements are less significant, as detailed in Tab.~\ref{tab:full results}. Furthermore, when applying the DINO discriminator to XL-XXL tokenizers, we observed that adversarial training frequently encounters instability. Specifically, a strong discriminator quickly learns to distinguish reconstructed samples, diminishing the benefits of adversarial training and leading to blurry artifacts. We leave further exploration of discriminator scaling and enhancement strategies for future work.

\begin{figure*}[]
% \vspace{-0.1in}
    \centering
    \includegraphics[width=\linewidth]{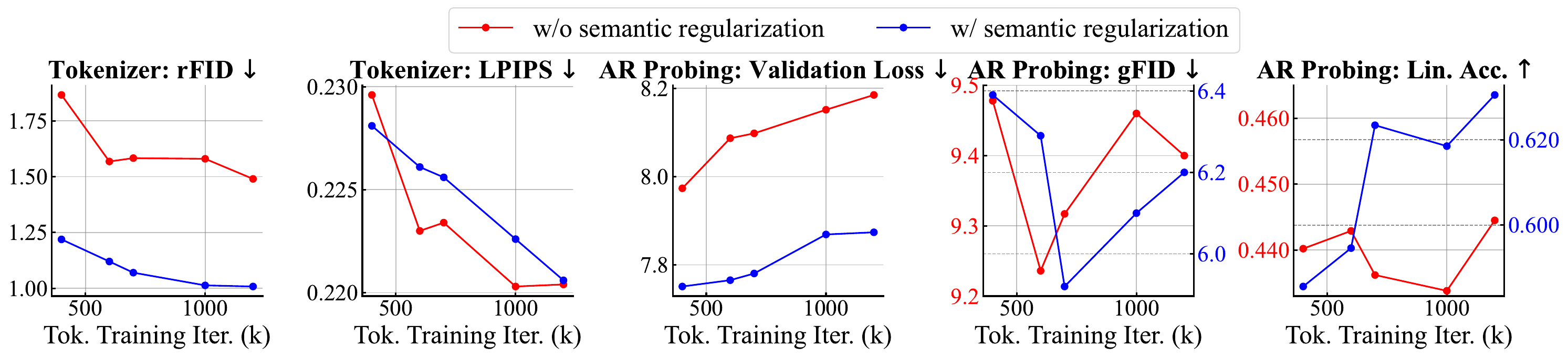}
    % \vspace{-0.30in}
    \caption{
    \textbf{Training duration scaling trends of tokenizers for reconstruction, downstream generation and representation quality with and without semantic regularization.} Note that in the last two figures, the red and blue curves correspond to different scales on the y-axis.
    }
    % \vspace{-0.3in}
    \label{fig:tok_iter_scale_cmp}
\end{figure*}

\section{Training Tokenizers for More Iterations}

While we largely resolve the reconstruction \vs generation dilemma regarding tokenizer \textbf{model scaling}, this challenge persists for tokenizer \textbf{training duration scaling}. To illustrate this phenomenon, we train five S-S tokenizers ranging from 40 to 120 epochs using a cosine learning rate scheduler, as detailed in Tab.~\ref{tab:training_setting}. The results are presented in Fig.~\ref{fig:tok_iter_scale_cmp}.

When extending tokenizer training iterations, reconstruction quality consistently improves. However, downstream generation quality initially improves but subsequently degrades with further increases in tokenizer training duration. Additionally, the validation loss of AR probing continuously rises with longer tokenizer training, regardless of semantic regularization. This trend suggests an increasing complexity in the tokenizer's latent space as the training duration extends.

We hypothesize that data scaling may alleviate this issue, and leave it for future exploration. In practice, allocating computational resources toward model scaling rather than extended training duration may yield better tokenizer performance.

\begin{figure}[]
% \vspace{-0.1in}
    \centering
    \includegraphics[width=\linewidth]{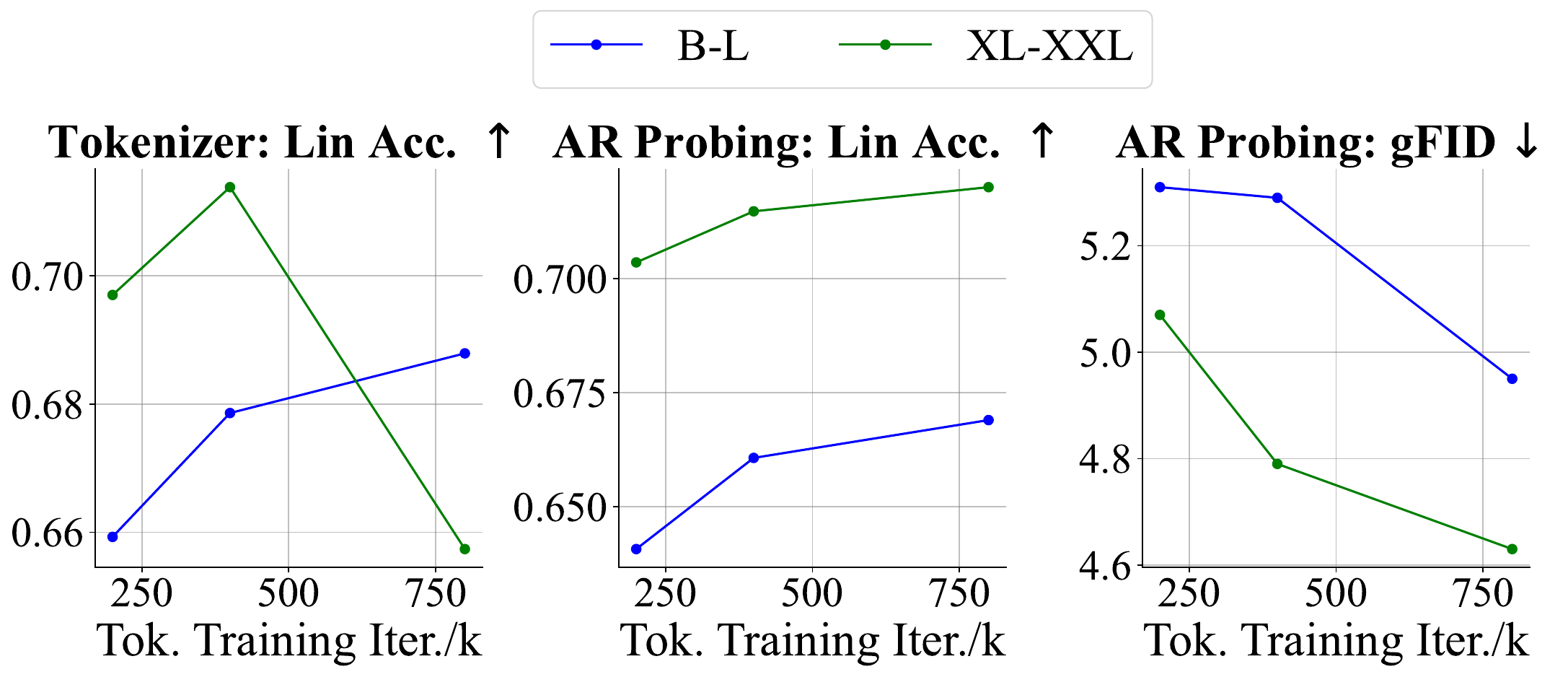}
    % \vspace{-0.30in}
    \caption{
    \textbf{The linear probing accuracy of tokenizer encoders does not necessarily reflect downstream model performance.} As the training proceeds, the XL-XXL tokenizer encoder presents an overfitting trend measured by linear probing accuracy, but downstream model performances consistently improve.
    }
    % \vspace{-0.3in}
    \label{fig:lin_acc_trend_tok_gpt}
\end{figure}

\section{Linear Probing Accuracy of Tokenizers}

We show that the linear probing accuracy of the tokenizer encoders may not necessarily indicate the performance of downstream AR models. We utilize the intermediate checkpoints during the training of B-L and XL-XXL tokenizers for evaluation. As shown in Fig.~\ref{fig:lin_acc_trend_tok_gpt}, the XL-XXL tokenizer encoder presents an overfitting trend in terms of tokenizer encoder linear probing accuracy. However, this overfitting trend is not reflected in AR Probing linear probing accuracy or gFID. Therefore, the linear probing accuracy of the tokenizer encoders may not be a good indicator of downstream model performance. Similarly, a concurrent work UniTok~\cite{ma2025unitok}, also points out that the performance of the tokenizer encoder in terms of zero-shot ImageNet classification accuracy may not necessarily reflect the visual understanding ability of downstream LLMs trained on the tokenizer.

The abnormality for large tokenizers reveals that the linear probing accuracy of the tokenizer is not necessarily a good indicator for downstream generation models. Since we care more about the representation learning for downstream models than for the tokenizers, using AR Probing as a direct evaluating method is better than indirect tokenizer linear probing accuracy.

\section{More Discussions About Related Work}
\noindent\textbf{TiTok}~\cite{titok} explores the use of 1D Transformer-based tokenizers under a high compression rate setting. TiTok seminally explores the model scaling of visual tokenizers and uses larger tokenizers for higher compression rate. 
However, the reconstruction \vs generation dilemma for scaling tokenizers is not solved in TiTok. As a result, the best generation model in TiTok is still trained on its smallest tokenizer variant. 

\noindent\textbf{ViTok}~\cite{scale_tokenizers} is a concurrent work which has explored the effect of model scaling for VAE~\cite{vae}. ViTok evaluates its VAE models in terms of both reconstruction and downstream diffusion generation performance. While having a very different setting from GigaTok, ViTok similarly finds that asymmetric design is better for VAEs. While ViTok suggests that small encoders are optimal, we point out that in our setting scaling encoders is also beneficial. Notably, the reconstruction \vs generation dilemma for scaling visual tokenizers is not solved in ViTok. We hypothesize that adding semantic regularization may similarly help solve the tokenizer scaling dilemma for VAEs, but leave it for future study.

\noindent\textbf{MAGVIT-v2}~\cite{magvit-v2} introduces LFQ to enhance discrete tokenizers. It also introduces the entropy penalty for tokenizer training, which is shown to be important for training large-scale tokenizers in our work. Instead of tokenizer model scaling, MAGVIT-v2 focuses more on scaling the codebook size of tokenizers. While codebook dimension and codebook size are important bottlenecks for visual tokenizers, we point out that model size scaling is also an important way for improving visual tokenizers.

\noindent\textbf{ImageFolder}~\cite{imagefolder} utilizes two branches for image encoding to handle high-level semantic information and low-level visual details respectively. It seminally utilizes semantic alignment to enhance the learned representation of tokenizers.

\noindent\textbf{VA-VAE}~\cite{lightningDiT} tames the reconstruction \vs generation dilemma in increasing latent dimensions for continuous VAE~\cite{vae, vae_journal}. VA-VAE improves the reconstruction-generation Pareto Frontier by introducing vision foundation model alignment loss. In contrast, we seek continuous improvements in both reconstruction and generation by scaling tokenizers. Semantic regularization serves different purposes in the two works.

\noindent\textbf{}

\end{document}